\pdfoutput=1

\documentclass[11pt]{article}

\usepackage[]{acl}

\usepackage{times}
\usepackage{latexsym}

\usepackage[T1]{fontenc}

\usepackage[utf8]{inputenc}

\usepackage{microtype}
\usepackage{graphicx, amsmath}
\title{VTCC-NLP at NL4Opt competition subtask 1: \\ An Ensemble Pre-trained language models for Named Entity Recognition}

\author{Xuan-Dung Doan \\
  Viettel Cyperspace Center, Viettel Group, Vietnam \\
    \texttt{dungdx4@viettel.com.vn} \\
}

\begin{document}
\maketitle
\begin{abstract}
We propose a combined three pre-trained language models (XLM-R, BART, and DeBERTa-V3) as an empower of contextualized embedding for named entity recognition. Our model achieves a 92.9\% F1 score on the test set and ranks 5th on the leaderboard at NL4Opt competition subtask 1 \cite{Ramamonjison22}.

\end{abstract}

\section{Introduction}
Named Entity Recognition (NER) is still a keyword for almost all researchers to find because of its effectiveness and importance in NLP problems. Particularly, NER includes three types: Flat NER, Discontinuous NER, and Nested NER. The NER problem in the NL4Opt competition is the Flat NER.

Recently, pre-trained language models have been applied in many problems of NLP.
XLM-R \cite{Conneau19} is a mask language model built based on RoBERTa \cite{Liu19} architecture. XLM-R was trained on the massive sizes of multilingual data.
BART \cite{Lewis19} is a denoising autoencoder for the sequence-to-sequence language model. BART is effective when fine-tuned for text generation task.
DeBERTa-V3 \cite{He21} replaced mask language modeling (MLM) with replaced token detection (RTD) and implemented a gradient-disentangled embedding sharing method. DeBERTa-V3 achieves SOTA on many downstream tasks of NLP.

Applying pre-trained language models boost the accuracy of low-resource and cross-domain tasks. In this regard, we propose an ensemble model from different pre-trained language
models for enriching semantic information of the embedding process.

\section{Related Works}
CRF \cite{Collobert11} is a popular technique used for NER. After that, many researchers use CRF with other methods such as \citet{Lample16}, \citet{Ma16}, \citet{Peters18}, \citet{Yue18}, \citet{Akbik19}.

Nowadays, pre-trained contextual embeddings such as BERT\cite{Devlin18}, BART, XLM-R, DeBERTA-V3, LUKE \cite{Ri22} have significantly improved the NER performance.

\section{Methodology}

\subsection{Encoder}
\begin{figure}[h!]
    \centering
    \captionsetup{justification=centering}
    \includegraphics[scale=0.29]{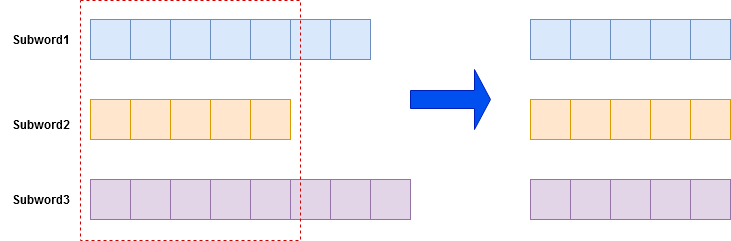}
    \caption{Subword selection}
    \label{Fig-1}
\end{figure}

Given input $X = \{x_1, x_2, ..., x_N\}$. The output is $Y=\{y_1, y_2, ..., y_N\}$ respectively. The output is a BIO format. This is a different tag is assigned to each word in the text depending on whether it is beginning (B-y), inside (I-y), or outside (O) a named entity phrase y.
Because different pre-trained language models give different subword tokenization. So, we select the one with the minimum length for representing the subword. Figure \ref{Fig-1} shows an example of subword selection.
The subword token was fed to pre-trained language models to get embedding information.

\begin{equation}
    V1 = BART(X)
\end{equation}
\begin{equation}
    V2 = XLMR(X)
\end{equation}
\begin{equation}
    V3 = DeBERTa\-V3(X)
\end{equation}

\begin{figure*}[h!]
\centering
\includegraphics[width=0.7\textwidth]{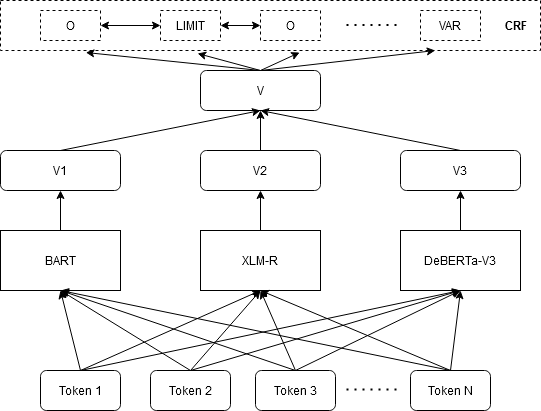}
\caption{Overview of the ensemble model.}
\label{Fig-2}
\end{figure*}

\subsection{Ensemble model}
\cite{wang22} use ensemble technique by majority voting m models with different seeds to get a final prediction. This method requires many models for prediction.
Therefore, we propose an ensemble pre-trained language model by
averaging their contextualized embedding to capture more
information. The combined embedding vector is calculated as
\begin{equation}
    V = \frac{\sum_{i=1}^{3} Vi}{3}
\end{equation}

\subsection{CRF Layer}
The token representations $V =\{v1, v2, ...v_n\}$ of the input sentence were put into a linear-chain CRF layer to obtain the conditional probability $p_{\theta}(y|\hat{x})$:
\begin{equation}
    \psi(y', y, v_i) = exp(W_y^T v_i + b_{y', y})
\end{equation}
\begin{equation}    
    p_{\theta}(y|\hat{x}) = \frac{\prod_{i=1}^{n} \psi(y', y, v_i)}{\sum_{y'\in Y(x)}\prod_{i=1}^{n} \psi(y', y, v_i)}
\end{equation}
where Y(x) denotes the set of all possible labels of the input
sequences x.

During training, we use the negative log-likelihood loss $L_{NLL}= -logp_\theta(y^*|\hat{x})$
For inference, the model predicts $\hat{y}_\theta$ by using the Viterbi algorithm.

Fig. \ref{Fig-2} illustrates the architecture of the proposed method

\section{Experiment}

\subsection{Dataset}

We use the official NL4Opt dataset. The training set, development set, and test set have 3 domains: sales, advertising, and investment. The test set expands 3 domains (unseen domains): transportation, production, and science.
The set of labels was: $LIMIT, CONST\_DIR, VAR, PARAM, \\ OBJ\_NAME, OBJ\_DIR$.

\subsection{Setting}
We set accumulating parameter to 4. The learning rate is 1e-4. The number of epochs was set to 50. We save the model after each epoch. The ten latest checkpoints are saving. We select the one with the best score on the dev set for the test set inference. All experiments are executed with a single 40G Nvidia-A100 GPU.

\subsection{Results}
\begin{figure}[h!]
\centering
\captionsetup{justification=centering}
\includegraphics[scale=0.6]{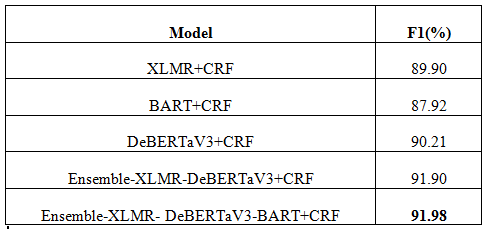}
\caption{Ensemble model on the development set.}
\label{Fig-3}
\end{figure}

The result on the dev set was shown in Fig. \ref{Fig-3}.
Our model is named "Ensemble-XLMR-DeBERTaV3-BART+CRF".

As shown in the reported table, the ensemble model method outperforms the single models. Our final ensemble models, including XLMR, DeBERTaV3, and BART,  give the best performance. 

\section{Dicussion}

\subsection{ELMo}
\begin{figure}[h!]
\centering
\captionsetup{justification=centering}
\includegraphics[scale=0.6]{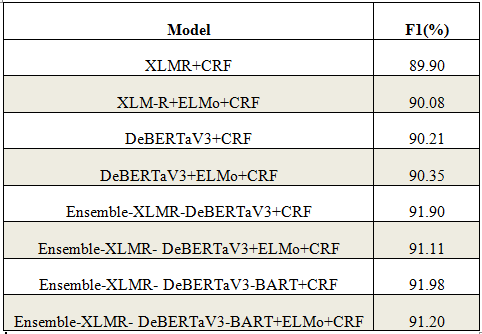}
\caption{Effectiveness of ELMo on the development set}
\label{Fig-4}
\end{figure}

To explore the contextualized embedding, we analyze the effect of ELMo embedding \cite{Peters18} on the dev set as shown in Fig. \ref{Fig-4}.

As result, using ELMo increases the accuracy of the baseline, but performance is lower for the ensemble model.

\subsection{GCN}
Graph neural networks were applied successfully in many problems of NLP. Inspired by \citet{jie19}, we implement the GCN model \cite{Kipf16}, which uses dependency parsing of each sentence as input. The result on the dev set was shown in Fig. \ref{Fig-5}:

\begin{figure}[h!]
\centering
\captionsetup{justification=centering}
\includegraphics[scale=0.6]{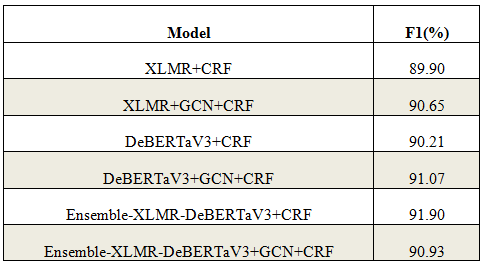}
\caption{Effectiveness of GCN on development set}
\label{Fig-5}
\end{figure}

Similarly ELMo, the results present that GCN increases the baseline model's accuracy but decreases the ensemble model's accuracy.

\bibliography{anthology,custom}

\begin{thebibliography}{16}
\expandafter\ifx\csname natexlab\endcsname\relax\def\natexlab#1{#1}\fi

\bibitem[{Akbik et~al.(2019)Akbik, Bergmann, Blythe, Rasul, Schweter, and
  Vollgraf}]{Akbik19}
Alan Akbik, Tanja Bergmann, Duncan Blythe, Kashif Rasul, Stefan Schweter, and
  Roland Vollgraf. 2019.
\newblock {FLAIR}: An easy-to-use framework for state-of-the-art {NLP}.
\newblock In \emph{{NAACL} 2019, 2019 Annual Conference of the North American
  Chapter of the Association for Computational Linguistics (Demonstrations)},
  pages 54--59.

\bibitem[{Collobert et~al.(2011)Collobert, Weston, Bottou, Karlen, Kavukcuoglu,
  and Kuksa}]{Collobert11}
Ronan Collobert, Jason Weston, L{\'{e}}on Bottou, Michael Karlen, Koray
  Kavukcuoglu, and Pavel~P. Kuksa. 2011.
\newblock \href {http://arxiv.org/abs/1103.0398} {Natural language processing
  (almost) from scratch}.
\newblock \emph{CoRR}, abs/1103.0398.

\bibitem[{Conneau et~al.(2019)Conneau, Khandelwal, Goyal, Chaudhary, Wenzek,
  Guzm{\'{a}}n, Grave, Ott, Zettlemoyer, and Stoyanov}]{Conneau19}
Alexis Conneau, Kartikay Khandelwal, Naman Goyal, Vishrav Chaudhary, Guillaume
  Wenzek, Francisco Guzm{\'{a}}n, Edouard Grave, Myle Ott, Luke Zettlemoyer,
  and Veselin Stoyanov. 2019.
\newblock \href {http://arxiv.org/abs/1911.02116} {Unsupervised cross-lingual
  representation learning at scale}.
\newblock \emph{CoRR}, abs/1911.02116.

\bibitem[{Devlin et~al.(2018)Devlin, Chang, Lee, and Toutanova}]{Devlin18}
Jacob Devlin, Ming{-}Wei Chang, Kenton Lee, and Kristina Toutanova. 2018.
\newblock \href {http://arxiv.org/abs/1810.04805} {{BERT:} pre-training of deep
  bidirectional transformers for language understanding}.
\newblock \emph{CoRR}, abs/1810.04805.

\bibitem[{He et~al.(2021)He, Gao, and Chen}]{He21}
Pengcheng He, Jianfeng Gao, and Weizhu Chen. 2021.
\newblock \href {http://arxiv.org/abs/2111.09543} {Debertav3: Improving deberta
  using electra-style pre-training with gradient-disentangled embedding
  sharing}.

\bibitem[{Jie and Lu(2019)}]{jie19}
Zhanming Jie and Wei Lu. 2019.
\newblock \href {https://doi.org/10.18653/v1/D19-1399} {Dependency-guided
  {LSTM}-{CRF} for named entity recognition}.
\newblock In \emph{Proceedings of the 2019 Conference on Empirical Methods in
  Natural Language Processing and the 9th International Joint Conference on
  Natural Language Processing (EMNLP-IJCNLP)}, pages 3862--3872, Hong Kong,
  China. Association for Computational Linguistics.

\bibitem[{Kipf and Welling(2016)}]{Kipf16}
Thomas~N. Kipf and Max Welling. 2016.
\newblock \href {http://arxiv.org/abs/1609.02907} {Semi-supervised
  classification with graph convolutional networks}.
\newblock \emph{CoRR}, abs/1609.02907.

\bibitem[{Lample et~al.(2016)Lample, Ballesteros, Subramanian, Kawakami, and
  Dyer}]{Lample16}
Guillaume Lample, Miguel Ballesteros, Sandeep Subramanian, Kazuya Kawakami, and
  Chris Dyer. 2016.
\newblock \href {http://arxiv.org/abs/1603.01360} {Neural architectures for
  named entity recognition}.
\newblock \emph{CoRR}, abs/1603.01360.

\bibitem[{Lewis et~al.(2019)Lewis, Liu, Goyal, Ghazvininejad, Mohamed, Levy,
  Stoyanov, and Zettlemoyer}]{Lewis19}
Mike Lewis, Yinhan Liu, Naman Goyal, Marjan Ghazvininejad, Abdelrahman Mohamed,
  Omer Levy, Veselin Stoyanov, and Luke Zettlemoyer. 2019.
\newblock \href {http://arxiv.org/abs/1910.13461} {{BART:} denoising
  sequence-to-sequence pre-training for natural language generation,
  translation, and comprehension}.
\newblock \emph{CoRR}, abs/1910.13461.

\bibitem[{Liu et~al.(2019)Liu, Ott, Goyal, Du, Joshi, Chen, Levy, Lewis,
  Zettlemoyer, and Stoyanov}]{Liu19}
Yinhan Liu, Myle Ott, Naman Goyal, Jingfei Du, Mandar Joshi, Danqi Chen, Omer
  Levy, Mike Lewis, Luke Zettlemoyer, and Veselin Stoyanov. 2019.
\newblock \href {http://arxiv.org/abs/1907.11692} {Roberta: {A} robustly
  optimized {BERT} pretraining approach}.
\newblock \emph{CoRR}, abs/1907.11692.

\bibitem[{Ma and Hovy(2016)}]{Ma16}
Xuezhe Ma and Eduard~H. Hovy. 2016.
\newblock \href {http://arxiv.org/abs/1603.01354} {End-to-end sequence labeling
  via bi-directional lstm-cnns-crf}.
\newblock \emph{CoRR}, abs/1603.01354.

\bibitem[{Peters et~al.(2018)Peters, Neumann, Iyyer, Gardner, Clark, Lee, and
  Zettlemoyer}]{Peters18}
Matthew~E. Peters, Mark Neumann, Mohit Iyyer, Matt Gardner, Christopher Clark,
  Kenton Lee, and Luke Zettlemoyer. 2018.
\newblock \href {http://arxiv.org/abs/1802.05365} {Deep contextualized word
  representations}.
\newblock \emph{CoRR}, abs/1802.05365.

\bibitem[{Ramamonjison et~al.(2022)Ramamonjison, Li, Yu, He, Rengan,
  Banitalebi-Dehkordi, Zhou, and Zhang}]{Ramamonjison22}
Rindranirina Ramamonjison, Haley Li, Timothy~T. Yu, Shiqi He, Vishnu Rengan,
  Amin Banitalebi-Dehkordi, Zirui Zhou, and Yong Zhang. 2022.
\newblock \href {https://doi.org/10.48550/ARXIV.2209.15565} {Augmenting
  operations research with auto-formulation of optimization models from problem
  descriptions}.

\bibitem[{Ri et~al.(2022)Ri, Yamada, and Tsuruoka}]{Ri22}
Ryokan Ri, Ikuya Yamada, and Yoshimasa Tsuruoka. 2022.
\newblock \href {https://aclanthology.org/2022.acl-long.505} {m{LUKE}: {T}he
  power of entity representations in multilingual pretrained language models}.
\newblock In \emph{Proceedings of the 60th Annual Meeting of the Association
  for Computational Linguistics (Volume 1: Long Papers)}. Association for
  Computational Linguistics.

\bibitem[{Wang et~al.(2022)Wang, Shen, Cai, Wang, Wang, Xie, Huang, Lu, Zhuang,
  Tu, Lu, and Jiang}]{wang22}
Xinyu Wang, Yongliang Shen, Jiong Cai, Tao Wang, Xiaobin Wang, Pengjun Xie, Fei
  Huang, Weiming Lu, Yueting Zhuang, Kewei Tu, Wei Lu, and Yong Jiang. 2022.
\newblock \href {https://doi.org/10.18653/v1/2022.semeval-1.200} {{DAMO}-{NLP}
  at {S}em{E}val-2022 task 11: A knowledge-based system for multilingual named
  entity recognition}.
\newblock In \emph{Proceedings of the 16th International Workshop on Semantic
  Evaluation (SemEval-2022)}, pages 1457--1468, Seattle, United States.
  Association for Computational Linguistics.

\bibitem[{Yang and Zhang(2018)}]{Yue18}
Jie Yang and Yue Zhang. 2018.
\newblock \href {https://doi.org/10.18653/v1/P18-4013} {{NCRF}++: An
  open-source neural sequence labeling toolkit}.
\newblock In \emph{Proceedings of {ACL} 2018, System Demonstrations}, pages
  74--79, Melbourne, Australia. Association for Computational Linguistics.

\end{thebibliography}
\bibliographystyle{acl_natbib}

\appendix

\end{document}